\documentclass[10pt,twocolumn,letterpaper]{article}

\usepackage{iccv}
\usepackage{times}
\usepackage{epsfig}
\usepackage{graphicx}
\usepackage{amsmath}
\usepackage{amssymb}
\usepackage{array}

\usepackage[pagebackref=true,breaklinks=true,letterpaper=true,colorlinks,bookmarks=false]{hyperref}

\iccvfinalcopy 

\def\iccvPaperID{367} 

\def\utc{\mbox{$\;\overset{c}{=}\;$}}

\ificcvfinal\fi
\begin{document}

\title{Volumetric Bias in Segmentation and Reconstruction: Secrets and Solutions}

\author{Yuri Boykov  \hspace{6ex} Hossam Isack   \\[1ex]
Computer Science \\
UWO, Canada \\
{\tt\small yuri@csd.uwo.ca \hspace{1ex}  habdelka@csd.uwo.ca }
\and
Carl Olsson\\[1ex]
Centre for Math. Sciences \\
Lund University, Sweden\\
{\tt\small calle@maths.lth.se}
\and
Ismail Ben Ayed\\[1ex]
Medical Biophysics\\
UWO, Canada\\
{\tt\small ibenayed@uwo.ca}
}

\pagestyle{myheadings}
\setlength{\headsep}{0.5in}
\markright{\small Y. Boykov, H. Isack, C. Olsson, I.B. Ayed,  arXiv:xxxx, May 2015  \hfill p.}
\setcounter{page}{1} 

\maketitle
\thispagestyle{myheadings}

\begin{abstract}
Many standard optimization methods for segmentation and reconstruction compute ML model 
estimates for appearance or geometry of segments, e.g. Zhu-Yuille \cite{Zhu:96}, Torr \cite{torr:98}, 
Chan-Vese \cite{Chan-Vese-01b}, GrabCut \cite{GrabCuts:SIGGRAPH04}, 
Delong et al. \cite{LabelCosts:IJCV12}.  We observe that the standard likelihood term 
in these formulations corresponds to a generalized probabilistic K-means energy. In learning it is well known 
that this energy has a strong bias to clusters of equal size, which can be expressed as a penalty 
for KL divergence from a uniform distribution of cardinalities \cite{UAI:97}. 
However, this volumetric bias has been mostly ignored in 
computer vision. We demonstrate significant artifacts in standard segmentation and reconstruction methods due to this bias. Moreover, we propose binary and multi-label optimization techniques that either (a) remove this bias or (b) replace it by a KL divergence term for 
any given target volume distribution. Our general ideas apply to many continuous or discrete
energy formulations in segmentation, stereo, and other reconstruction problems.
\end{abstract}

\section{Introduction}

Most problems in computer vision are ill-posed and optimization of regularization functionals
is critical for the area. In the last decades the community developed many practical energy functionals and
efficient methods for optimizing them. This paper analyses a widely used general class of segmentation
energies motivated by Bayesian analysis, discrete graphical models (e.g. MRF/CRF), information theory (e.g. MDL) , 
or continuous geometric formulations. Typical examples in this class of energies 
include a log-likelihood term for models $P^k$ assigned to image segments $S^k$
\begin{equation} \label{eq:lk_term}
E(S,P) = - \sum_{k=1}^{K} \sum_{p\in S^k}  \log P^k(I_p),
\end{equation}
where, for simplicity, we focus on a discrete formulation with data $I_p$ for a finite set of pixels/features $p\in\Omega$ and 
segments $S^k = \{p\in\Omega | S_p=k\}$ defined by variables/labels $S_p \in {\bf N}$ 
indicating the segment index assigned to $p$. In different vision problems models $P^k$ 
could represent Gaussian intensity models \cite{Chan-Vese-01b}, color histograms \cite{BJ:01},
GMM \cite{Zhu:96,GrabCuts:SIGGRAPH04}, or geometric models \cite{torr:98,LabelCosts:IJCV12,Barinova:cvpr10}  
like lines, planes, homographies, or fundamental matrices.
\begin{figure}[t]
\begin{tabular}{c@{\extracolsep{1mm}}c}
{\bf Secrets} \eqref{eq:lk_term} & {\bf Solutions} \eqref{eq:wlk_term},  (\ref{eq:E2}-\ref{eq:E2_H}) \\
\includegraphics[width=38mm]{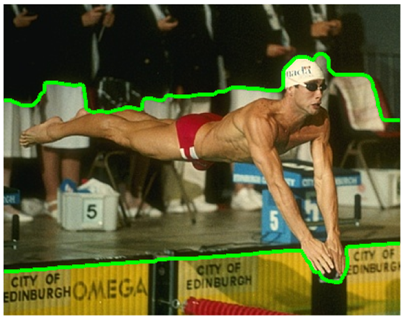} &
\includegraphics[width=38mm]{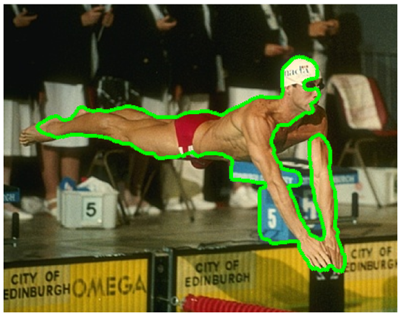} \\[-0.9ex]
(a) GrabCut \cite{GrabCuts:SIGGRAPH04} & with unbiased data term \eqref{eq:E2_H} \\[0.5ex]
\includegraphics[width=38mm]{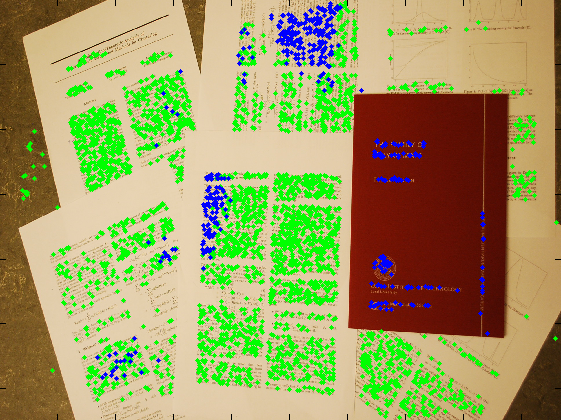} &
\includegraphics[width=38mm]{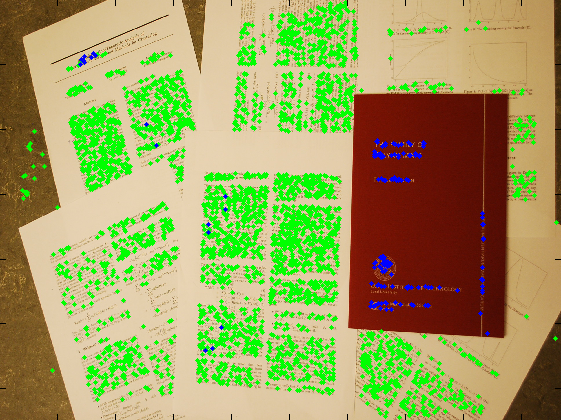} \\[-0.9ex]
(b) plane fitting \cite{torr:98,LabelCosts:IJCV12,Barinova:cvpr10} & with unbiased data term \eqref{eq:E2_H} \\[0.5ex]
\includegraphics[width=38mm]{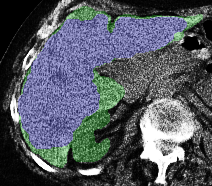} &
\includegraphics[width=38mm]{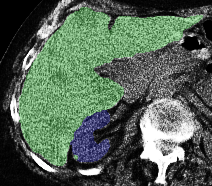} \\[-0.9ex]
(c) Chan-Vese \cite{Chan-Vese-01b} $+$ \cite{DB:ICCV09}   &  with target volumes \eqref{eq:wlk_term}  \\[1ex]
\end{tabular}
\caption{\emph{Left:} segmentation and stereo reconstruction with standard likelihoods or {\em probabilistic K-means} energy $E(S,P)$ in \eqref{eq:lk_term} has
bias to equal size segments \eqref{eq:KL_U}. \emph{Right:} (a-b) corrections due to unbiased data term $\hat{E}(S,P)$ in (\ref{eq:E2},\ref{eq:E2_H})
or (c) weighted likelihoods  $E_W(S,P)$  in \eqref{eq:wlk_term} biased to proper target volumes, see \eqref{eq:KL_W}.
Sections \ref{sec:sgm_target}-\ref{sec:stereo} explain these examples in details. \label{fig:teaser}}
\end{figure}

Depending on application, the energies combine likelihoods \eqref{eq:lk_term}, a.k.a. data term, with different regularization potentials 
for segments $S^k$. One of the most standard regularizers is the Potts potential, as in the following energy
\begin{equation*}
E_{Potts}(S,P) = - \sum_{k=1}^{K} \sum_{p\in S^k} \log P^k(I_p)+\lambda\cdot ||\partial S||,
\end{equation*}
where $ ||\partial S||$ is the number of label discontinuities between neighboring points $p$ on
a given neighborhood graph or the length of the segmentation boundary in the image grid \cite{BK:iccv03}.
Another common regularizer is sparsity or label cost for each model $P^k$ with non-zero support 
\cite{torr:98,Zhu:96,Barinova:cvpr10,LabelCosts:IJCV12}, \eg
\begin{equation*}
E_{sp}(S,P) = - \sum_{k=1}^{K} \sum_{p\in S^k} \log P^k(I_p) + \gamma \cdot \sum_k [S^k\neq \emptyset].
\end{equation*}
In general, energies often combine likelihoods \eqref{eq:lk_term} with multiple different regularizers
at the same time.

This paper demonstrates practically significant bias to equal size segments in standard 
energies when models $P=\{P^k\}$ are treated as variables jointly estimated with segmentation $S=\{S^k\}$.
This problem comes from likelihood term \eqref{eq:lk_term}, which we interpret as 
{\em probabilistic K-means} energy carefully analyzed in \cite{UAI:97} from an information theoretic point of view.
In particular, \cite{UAI:97} decomposes energy \eqref{eq:lk_term} 
as\footnote{Symbol $\utc$ represents equality up to an additive constant.}
\begin{equation*}
E(S,P) \utc \sum_{k=1}^{K} |S^k|\cdot KL(I^k|P^k) + |\Omega|\cdot ( H(S|I) - H(S))
\end{equation*}
where $KL(I^k|P^k)$ is KL divergence for model  $P^k$ and the true 
distribution\footnote{The decomposition above applies to either discrete or continuous probability models 
(e.g. histogram vs. Gaussian). The continuous case relies on Monte-Carlo estimation 
of the integrals over ``true'' data density. } 
of data $I^k=\{I_p\;|\;p\in S^k\}$ in segment $k$. Conditional entropy $H(S|I)$ penalizes ``non-deterministic'' 
segmentation if variables $S_p$ are not completely determined by intensities $I_p$. The last term is
negative entropy of segmentation variables $-H(S)$, which can be seen as KL divergence
\begin{equation} \label{eq:KL_U}
-H(S) \;\utc\; KL(S|U) \;:=\; \sum_{k=1}^{K} \frac{|S^k|}{|\Omega|}\cdot\ln\frac{|S^k|/|\Omega|}{1/K}
\end{equation}
between the volume distribution for segmentation $S$
\begin{equation} \label{eq:VS}
V_S := \left\{\frac{|S^1|}{|\Omega|},\frac{|S^2|}{|\Omega|},\dots,\frac{|S^K|}{|\Omega|}\right\}
\end{equation}
and a uniform distribution $U=\{\frac{1}{K},...,\frac{1}{K}\}$. Thus, this term represents volumetric bias 
to equal size segments $S^k$. Its minimum is achieved for cardinalities $|S^k| = \frac{\Omega}{K}$.

\subsection{Contributions} \label{sec:intro2}
Our experiments demonstrate that volumetric bias in probabilistic K-means energy \eqref{eq:lk_term}
leads to practically significant artifacts for problems in computer vision, where this term is widely 
used for model fitting in combination with different regularizers, e.g. 
\cite{Zhu:96,torr:98,Chan-Vese-01b,GrabCuts:SIGGRAPH04,LabelCosts:IJCV12}.
Section \ref{seg:energy_formulation} proposes several ways to address this bias.

First, we show how to remove the volumetric bias. This could be achieved by adding extra term
$|\Omega|\cdot H(S)$ to any energy with likelihoods \eqref{eq:lk_term} exactly compensating for 
the bias. We discuss several efficient optimization techniques applicable to this high-order energy term in continuous
and/or discrete formulations: iterative bound optimization, exact optimization for binary discrete problems, and
approximate optimization for multi-label problems using $\alpha$-expansion  \cite{BVZ:pami01}.
It is not too surprising that there are efficient solvers for the proposed correction term 
since $H(S)$ is a concave cardinality function, which is known to be {\em submodular} 
for binary problems \cite{lovasz:83}. 
Such terms have been addressed previously, in a different context, in the vision literature \cite{kohli:09,onecut:13}.

Second, we show that the volumetric bias to uniform distribution could be replaced by a bias
to any given target distribution of cardinalities 
\begin{equation} \label{eq:W}
W=\{w^1,w^2,...,w^K\}.
\end{equation}
In particular, introducing weights $w^k$ for log-likelihoods in \eqref{eq:lk_term} replaces 
bias $KL(S|U)$ as in \eqref{eq:KL_U} by divergence between segment volumes and desired target distribution $W$
\begin{equation} \label{eq:KL_W}
 KL(S|W) = \sum_{k=1}^{K} \frac{|S^k|}{|\Omega|}\cdot\ln\frac{|S^k|/|\Omega|}{w^k}. 
\end{equation}

Our experiments in supervised or unsupervised segmentation and in stereo reconstruction 
demonstrate that both approaches to managing volumetric bias in \eqref{eq:lk_term}
can significantly improve the robustness of many energy-based methods for computer vision.


\section{Log-likelihood energy formulations}\label{seg:energy_formulation}

This section has two goals. First, we present weighted likelihood energy $E_W(S,P)$ in \eqref{eq:wlk_term} 
and show in \eqref{eq:EW_KL} that its volumetric bias is defined by $KL(S|W)$. Standard data term $E(P,S)$ in \eqref{eq:lk_term}
is a special case with $W=U$. Then, we present another modification of likelihood energy $\hat{E}(S,P)$ in \eqref{eq:E2} and 
prove that it does not have volumetric bias.  Note that \cite{UAI:97} also discussed unbiased energy $\hat{E}$. 
The analysis of $\hat{E}$  below is needed for completeness and to devise optimization for problems in vision 
where likelihoods are only a part of the objective function.

{\bf Weighted likelihoods:} Consider energy
\begin{equation} \label{eq:wlk_term}
E_W(S,P) := - \sum_{k=1}^{K} \sum_{p\in S^k}  \log ( w^k \cdot P^k(I_p)),
\end{equation}
which could be motivated by a Bayesian interpretation \cite{LabelCosts:IJCV12} 
where weights $W$ explicitly come from a volumetric prior.
It is easy to see that
\begin{eqnarray} \label{eq:EW}
E_W(S,P) & = & E(S,P) - \sum_{k=1}^{K} |S^k|\cdot \log w^k \\
               & = & E(S,P) + |\Omega|\cdot H(S|W)   \nonumber
\end{eqnarray}
where $H(S|W)$ is a cross entropy between distributions $V_S$ and $W$.
As discussed in the introduction, the analysis of probabilistic K-means energy $E(S,P)$ in \cite{UAI:97} implies that
\begin{eqnarray*}
E_W(S,P)  & \utc & \sum_{k=1}^{K} |S^k|\cdot KL(I^k|P^k) + |\Omega|\cdot H(S|I) \\
                 & - & |\Omega|\cdot H(S) + |\Omega|\cdot H(S|W).
\end{eqnarray*}
Combining two terms in the second line gives
\begin{eqnarray} 
E_W(S,P)  & \utc & \sum_{k=1}^{K} |S^k|\cdot KL(I^k|P^k) + |\Omega|\cdot H(S|I)   \nonumber \\
                  &  +    & |\Omega|\cdot KL(S|W).          \label{eq:EW_KL}
\end{eqnarray}
In case of given weights $W$ equation \eqref{eq:EW_KL} implies that  weighted likelihood term \eqref{eq:wlk_term} 
has bias to the target volume distribution represented by KL divergence \eqref{eq:KL_W}. 

Note that optimization of weighted likelihood term \eqref{eq:wlk_term} presents no extra difficulty
for regularization methods in vision. Fixed weights $W$ contribute unary potentials for segmentation 
variables $S_p$, see \eqref{eq:EW}, which are trivial for standard discrete or
continuous optimization methods. Nevertheless, examples in Sec.~\ref{sec:examples} show that
indirect minimization of KL divergence \eqref{eq:KL_W}  substantially improves the results in applications
if (approximate) target volumes $W$ are known.

{\bf Unbiased data term:} If weights $W$ are treated as unknown parameters in likelihood energy \eqref{eq:wlk_term} 
they can be optimized out. In this case decomposition \eqref{eq:EW_KL} implies that the corresponding energy 
has no volumetric bias:
\begin{eqnarray}\label{eq:E2}
\hat{E}(S,P)&:=&\min_W E_W(S,P) \\
                   & = & \sum_{k=1}^{K} |S^k|\cdot KL(I^k|P^k) + |\Omega|\cdot H(S|I) .  \nonumber
\end{eqnarray}
Weights $V_S$ in \eqref{eq:VS} are ML estimate of $W$ that minimize \eqref{eq:EW_KL} by achieving $KL(S|W)=0$.
Putting optimal weights $W=V_S$ into \eqref{eq:EW} confirms that volumetrically unbiased data term \eqref{eq:E2}
is a combination of standard likelihoods \eqref{eq:lk_term} with a high-order correction term $H(S)$:
\begin{eqnarray}
\hat{E}(S,P) &=&  E(S,P) - \sum_{k=1}^{K} |S^k|\cdot \log \frac{|S^k|}{|\Omega|}  \nonumber \\
&=& E(S,P) + |\Omega|\cdot H(S). \label{eq:E2_H}
\end{eqnarray}

Note that unbiased data term $\hat{E}(S,P)$ should be used with caution in applications 
where allowed models $P^k$ are highly descriptive. In particular, this applies to Zhu\&Yuille \cite{Zhu:96}
and GrabCut \cite{GrabCuts:SIGGRAPH04} where probability models are histograms or GMM.
According to \eqref{eq:E2}, optimization of model $P^k$ will over-fit to data, \ie $KL(I^k|P^k)$ 
will be reduced to zero for arbitrary $I^k = \{I_p\;|\;p\in S^k\}$. Thus, highly descriptive models 
reduce $\hat{E}(S,P)$ to conditional entropy $H(S|I)$, which only encourages consistent labeling 
for points of the same color. While this could be useful in segmentation, see {\em bin consistency} 
in \cite{onecut:13}, trivial solution $S^0=\Omega$ becomes good for energy $\hat{E}(S,P)$. 
Thus, bias to equal size segments in standard likelihoods \eqref{eq:lk_term} is important 
for histogram or GMM fitting methods \cite{Zhu:96,GrabCuts:SIGGRAPH04}. 

Many techniques with unbiased data term $\hat{E}(S,P)$ avoid trivial solutions.
Over-fitting is not a problem for simple models, e.g. Gaussians \cite{Chan-Vese-01b}, 
lines, homographies \cite{torr:98,LabelCosts:IJCV12}. Label cost could be used 
to limit model complexity. Trivial solutions could also be removed by specialized
regional terms added to the energy \cite{onecut:13}. Indirectly, 
optimization methods that stop at a local minimum help as well. 

\begin{figure}[t]
\begin{center}
\includegraphics[width=0.85\linewidth]{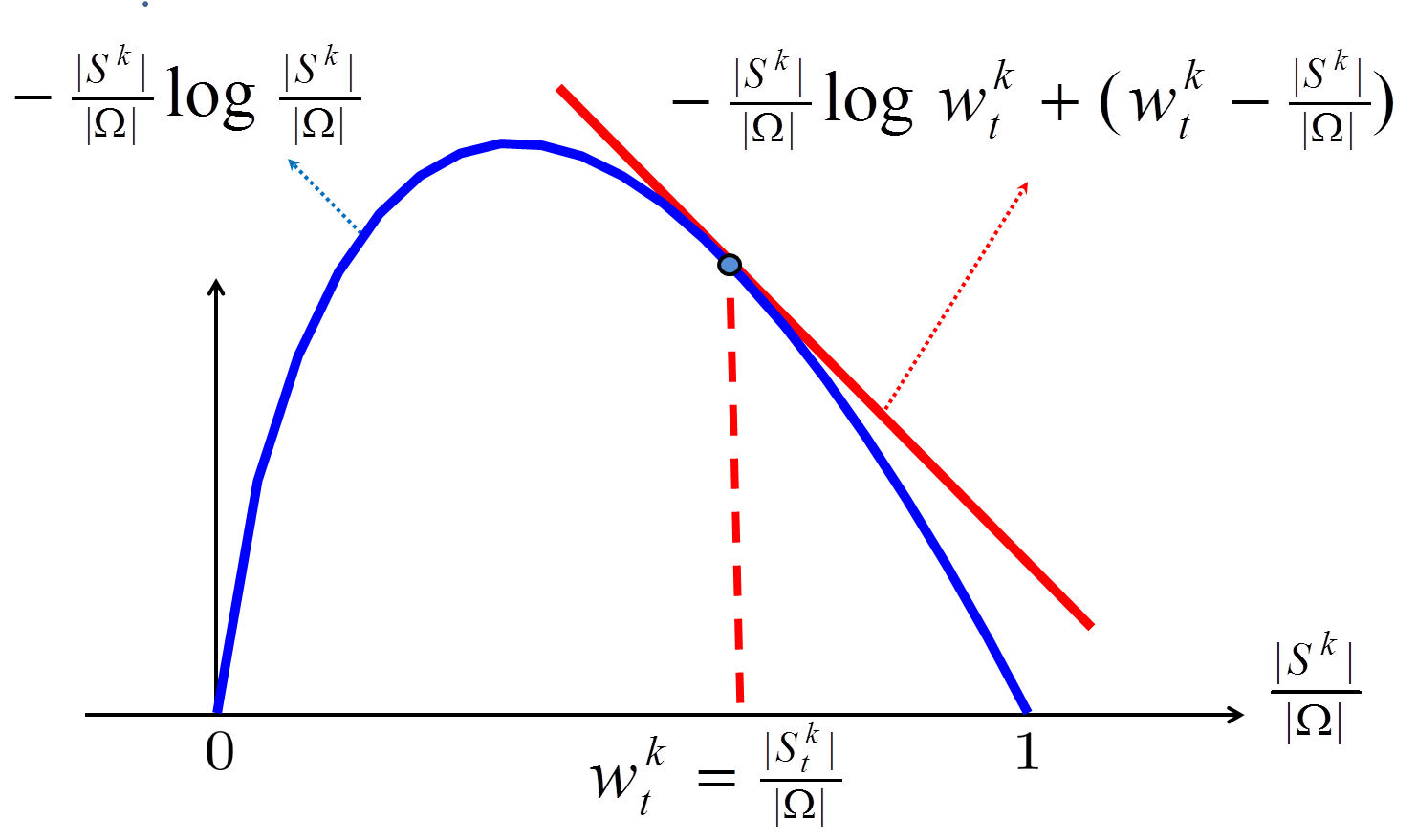} 
\caption{({\em Entropy - bound optimization}) According to (\ref{eq:EW},\ref{eq:E2_H}) energy $E_{W_t}(S,P)$ is a bound for $\hat{E}(S,P)$ since cross entropy $H(S|W_t)$ is a bound for entropy $H(S)$ with equality at $S=S_t$. 
This standard fact is easy to check: function $-z \log z$ (blue curve) is concave and 
its 1st-order approximation at $z_t = w_t^k = |S^k_t|/|\Omega|$ (red line) 
is a tight upper-bound or {\em surrogate function} \cite{Lange2000}.
\label{fig:bound}}
\end{center} \vspace{-2ex}
\end{figure}
\begin{figure}
\begin{center} 
\includegraphics[width=40mm]{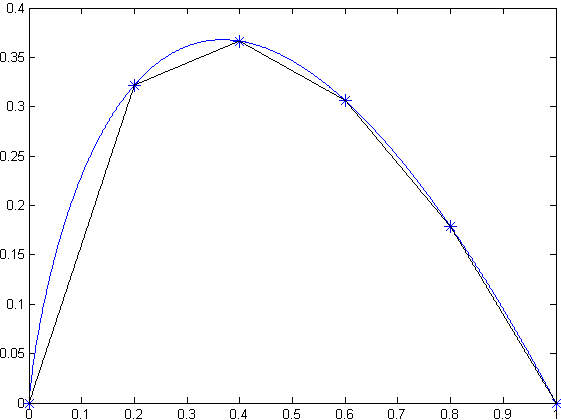}
\includegraphics[width=40mm]{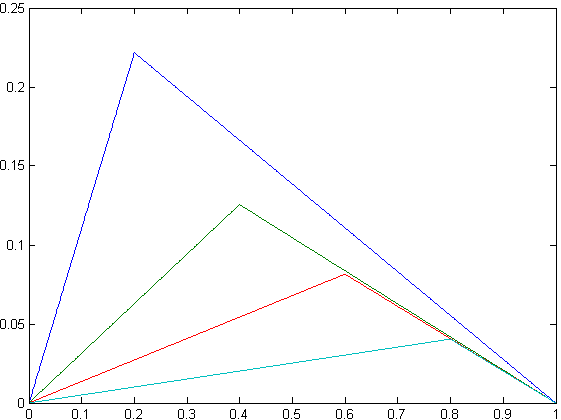}
\caption{({\em Entropy - high order optimization}) (a) polygonal approximation for $-z \log z$. 
(b) ``triangle'' functions decomposition.
\label{fig:entropyapprox}}
\end{center} \vspace{-5ex}
\end{figure}

{\bf Bound optimization for (\ref{eq:E2}-\ref{eq:E2_H}):} 
One local optimization approach for $\hat{E}(S,P)$ uses iterative minimization of weights $W$ for $E_W(S,P)$. 
According to \eqref{eq:EW_KL} the optimal weights at any current solution $S_t$ are
$W_t=\{\frac{|S^1_t|}{|\Omega|},...,\frac{|S^K_t|}{|\Omega|}\}$ since they minimize $KL(S_t|W)$.
The algorithm iteratively optimizes $E_{W_t}(S,P)$ over $P, S$ and resets to energy $E_{W_{t+1}}(S,P)$ at each 
step until convergence. This block-coordinate descent can be seen as {\em bound optimization} \cite{Lange2000}. 
Indeed, see Figure \ref{fig:bound}, at any given $S_t$ energy $E_{W_t}(S,P)$ 
is an {\em upper bound} for $\hat{E}(S,P)$, that is
\begin{eqnarray*}
\hat{E}(S,P)    & \leq &  E_{W_t}(S,P)\;\;\;\;\;\;\forall S \\
\hat{E}(S_t,P) &  =   &  E_{W_t}(S_t,P).
\end{eqnarray*}
This bound optimization approach to $\hat{E}(S,P)$ is a trivial modification for any standard optimization
algorithm for energies with unary likelihood term $E_W(S,P)$ in \eqref{eq:wlk_term}.

{\bf High-order optimization for entropy in (\ref{eq:E2}-\ref{eq:E2_H}):} 
Alternatively,  optimization of unbiased term $\hat{E}(S,P)$ could be based on equation 
\eqref{eq:E2_H}. Since term $E(S,P)$ is unary for $S$ the only issue is optimization of high-order
entropy $H(S)$. The entropy is a combination of terms $-z \log z$ for $z = |S^k|/|\Omega|$. 
Each of these is a concave function of cardinality, which are known to be {\em submodular} \cite{lovasz:83}. 
As explained below, entropy is amenable to efficient discrete optimization 
techniques both in binary (Sec.\ref{sec:sgm_no_bias}) and multi-label cases 
(Sec.\ref{sec:sgm_no_bias}-\ref{sec:stereo}).

Optimization of concave cardinality functions was previously proposed in vision
for {\em label consistency}  \cite{kohli:09}, {\em bin consistency}  \cite{onecut:13}, and other applications. 
Below, we discuss similar optimization methods in the context of entropy.
We use a polygonal approximation with triangle functions as illustrated in Figure~\ref{fig:entropyapprox}. 
Each triangle function is the minimum of two affine cardinality functions, yielding an approximation of the type
\begin{equation}
-\frac{|S^k|}{|\Omega|}\log\frac{|S^k|}{|\Omega|} \approx\sum_l \min\left(a_l^L |S^k|, a_l^U |S^k|+b_l^U\right).
\end{equation}
Optimization of each ``triangle'' term in this summation can be done as follows.
Cardinality functions like $a_l^L |S^k|$ and $a_l^U |S^k|+b_l^U$ are unary. 
Evaluation of their minimum can be done with an {\em auxiliary} variable $y_l\in\{0,1\}$ as in
\begin{equation} 
\min_{y_l} y_l (a_l^L |S^k|) + \bar{y}_l (a_l^U |S^k|+b_l^U)  \label{eq:linE}
\end{equation}
which is a pairwise energy. Indeed, consider binary segmentation problems $S_p\in\{0,1\}$. Since
\begin{equation}
|S^k| = 
\begin{cases} \sum_{p\in\Omega} S_p, & \mbox{if } k=1  \\
                         \sum_{p\in\Omega} (1-S_p), & \mbox{if } k=0
\end{cases} 
\end{equation}
\eqref{eq:linE} breaks into  {\em submodular}\footnote{Depending on $k$, may need to switch $y_l$ and $\bar{y}_l$.}
pairwise terms for $y_l$ and $S_p$. Thus, each ``triangle'' energy \eqref{eq:linE} can be globally optimized with 
graph cuts \cite{Vlad:PAMI04}.
For more general multi-label problems $S_p\in N$ energy terms \eqref{eq:linE} can be iteratively optimized via binary graph-cut 
moves like $\alpha$-expansion \cite{BVZ:pami01}. Indeed, let variables $x_p\in\{0,1\}$ represent $\alpha$-expansion from a current solution $S_t = \{S^k_t\}$ 
to a new solution $S$. Since
\begin{equation}
|S^k| = 
\begin{cases} \sum_{p\in\Omega} x_p, & \mbox{if } k=\alpha  \\
                         \sum_{p\in S^k_t} (1-x_p), & \mbox{if } k\neq \alpha
\end{cases} 
\end{equation}
\eqref{eq:linE} also reduces to submodular pairwise terms for $y_l$, $x_p$.

The presented high-order optimization approach makes stronger moves than the simpler bound optimization
method in the previous sub-section. However, both methods use block coordinate descent iterating 
optimization of $S$ and $P$ with no quality guarantees. 
The next section shows examples with different optimization methods.

\begin{figure*}[th]
\begin{center}
\begin{tabular}{c@{\extracolsep{2mm}}c@{\extracolsep{2mm}}c@{\extracolsep{2mm}}c}
(a) initial models  & (b) segmentation for $W=U$ & 
(c) for $W=\{0.04,0.96 \}$ & (d) for $W=\{0.75,0.25 \}$\\
\includegraphics[width=40mm]{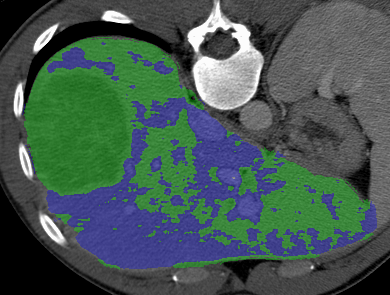} &
\includegraphics[width=40mm]{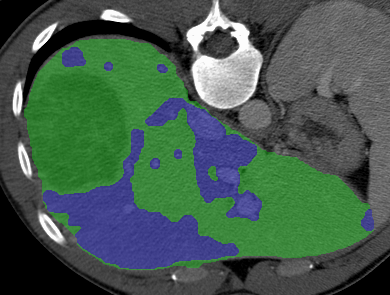} &
\includegraphics[width=40mm]{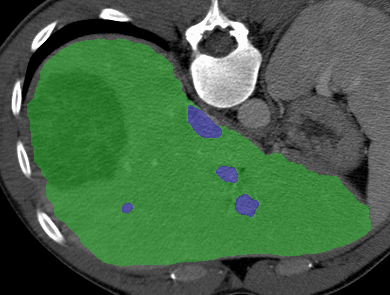} &
\includegraphics[width=40mm]{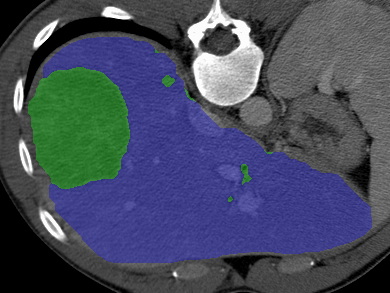} \\
\includegraphics[width=40mm]{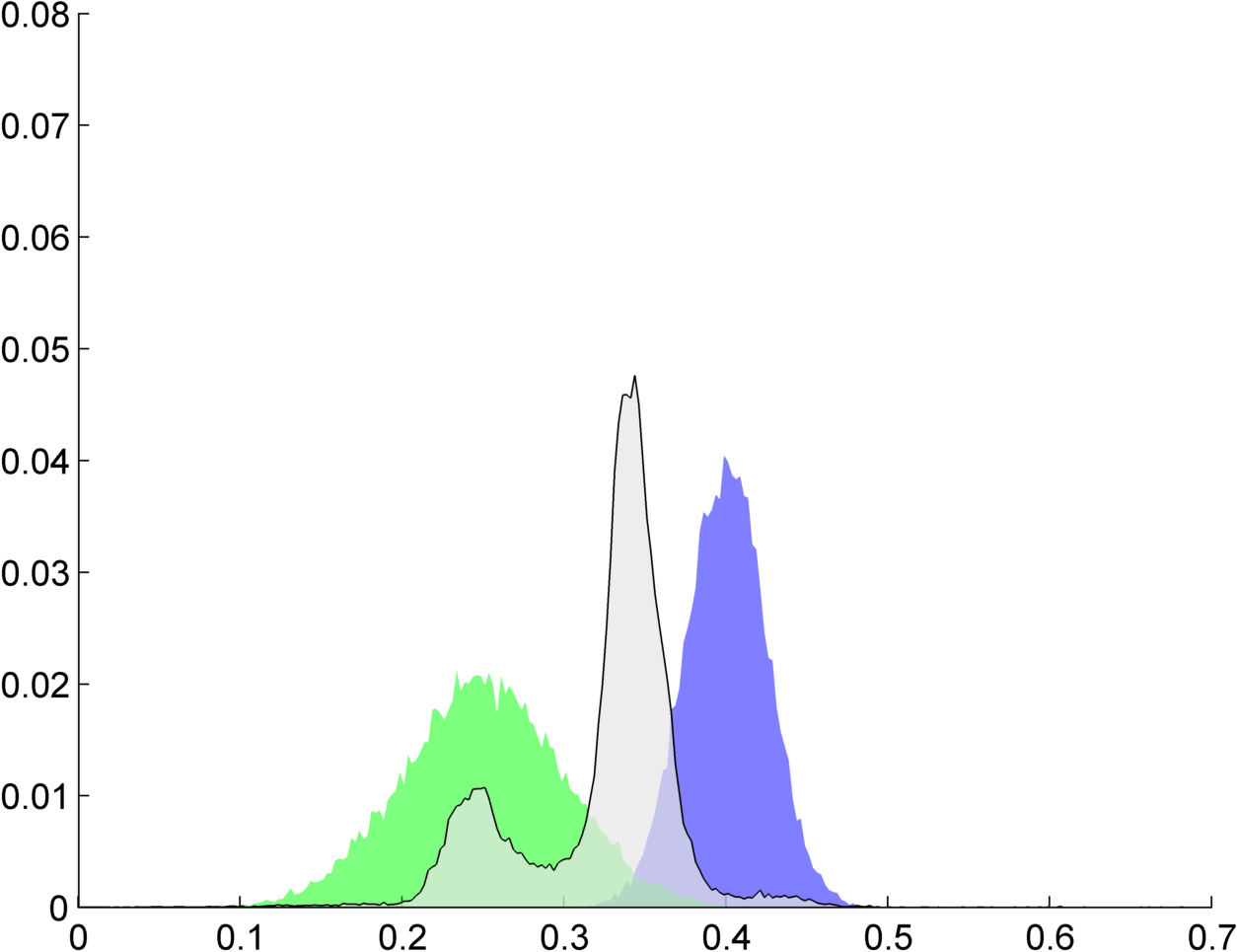} &
\includegraphics[width=40mm]{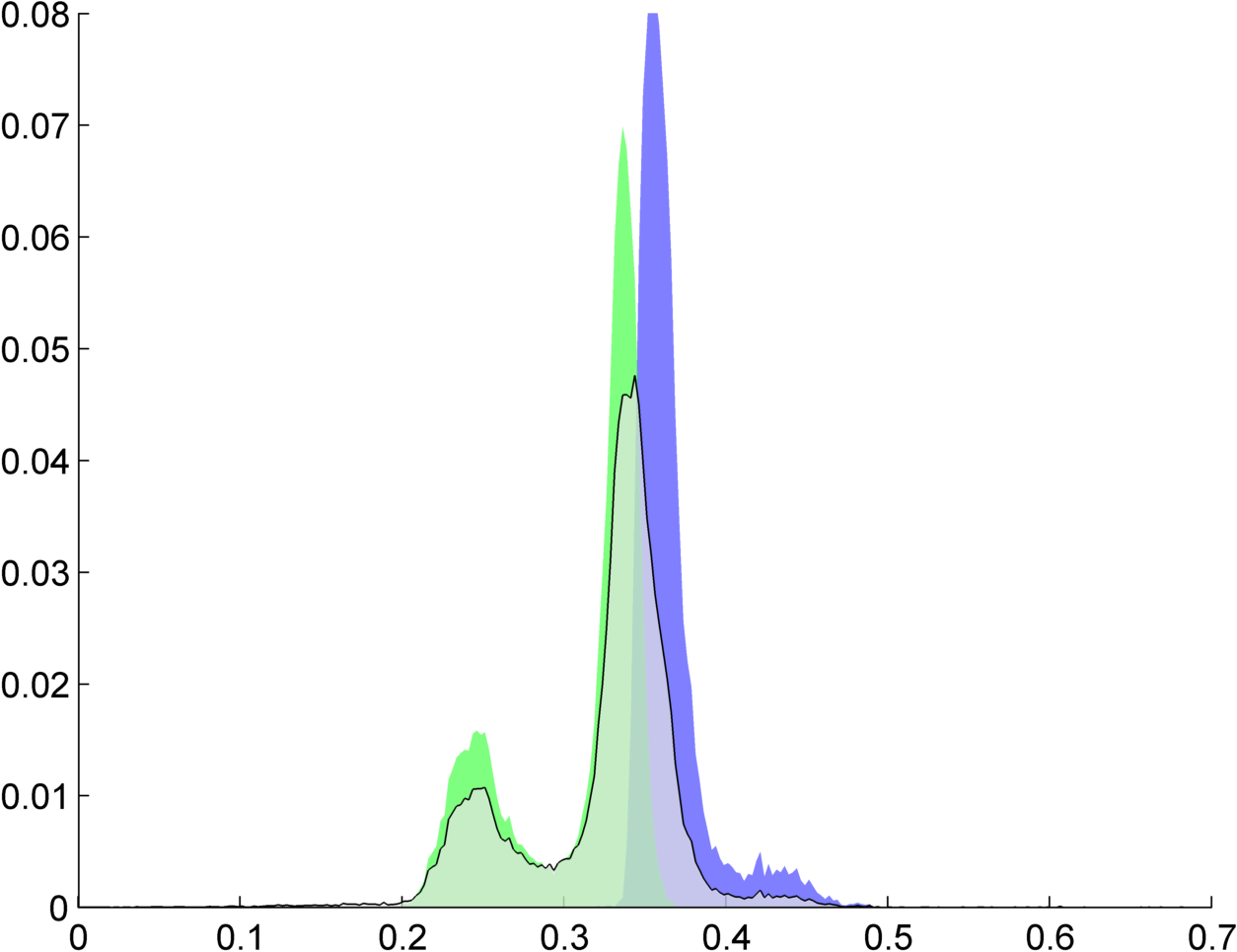} &
\includegraphics[width=40mm]{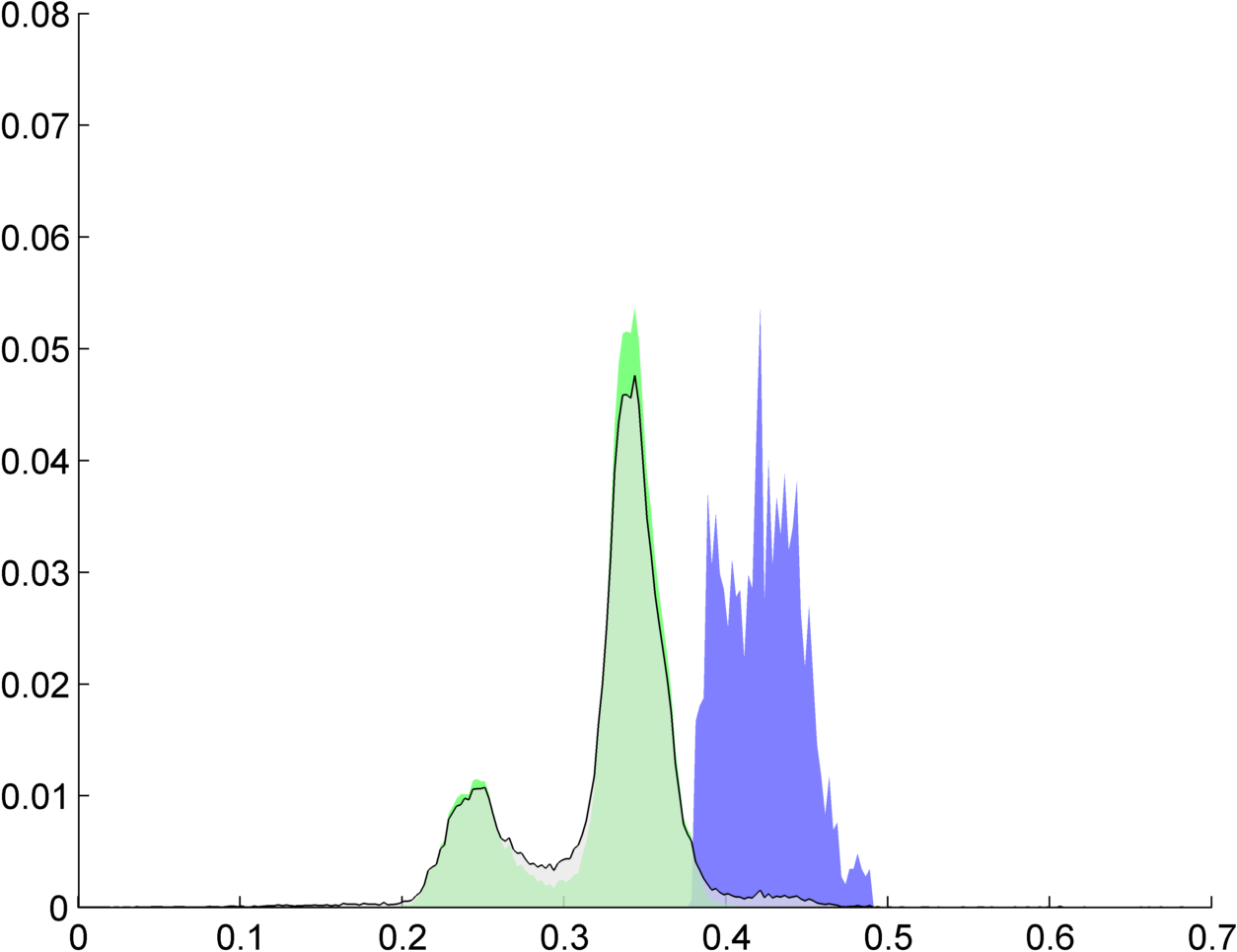} &
\includegraphics[width=40mm]{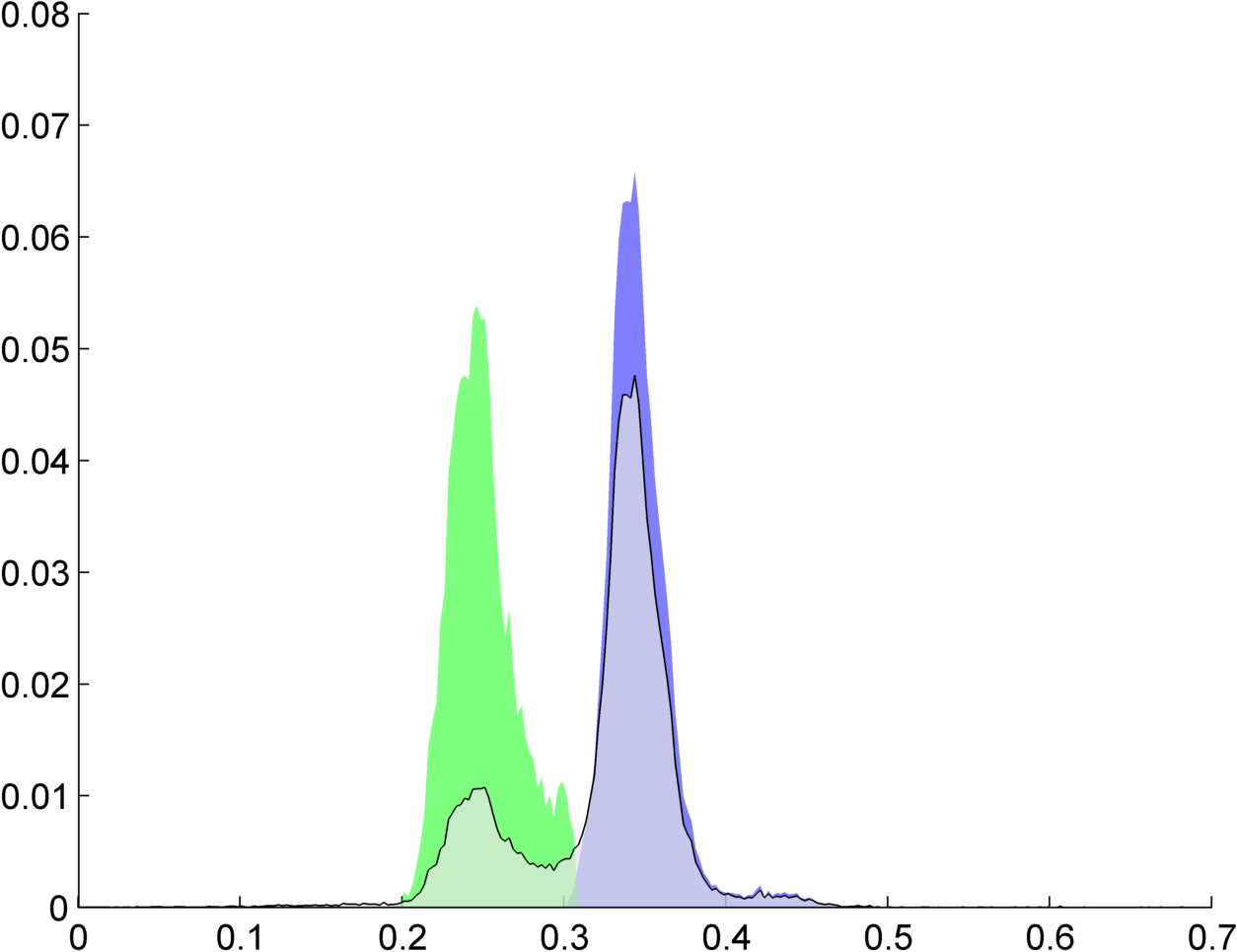} \\[-1ex]
\end{tabular}
\end{center}
\caption{Equal volumes bias $KL(S|U)$ versus target volumes bias $KL(S|W)$. Grey histogram is a distribution 
of intensities for the ground truth liver segment including normal liver tissue (the main mode), blood vessels 
(the small mode on the right), and cancer tissue (the left mode). 
(a) Initial (normalized) histograms for two liver parts. Initial segmentation shows which histogram 
has larger value for each pixel's intensity. (b) The result of optimizing energy \eqref{eq:EDB1}. 
The solid blue and green histograms at the bottom row are for intensities at
the corresponding segments. (c-d) The results of optimizing energy \eqref{eq:EDB2} for fixed weights $W$ set for
specific target volumes. }
\label{fig:target}
\end{figure*}

\section{Examples} \label{sec:examples}
This sections considers several representative examples of computer vision problems 
where regularization energy uses likelihood term \eqref{eq:lk_term} with re-estimated models $P^k$.
We empirically demonstrate bias to segments of the same size \eqref{eq:KL_U} 
and show advantages of different modifications of the data term proposed
in the previous section.

\subsection{Segmentation with target volumes} \label{sec:sgm_target}
In this section we consider a biomedical example with
 $K=3$ segments: $S_1$ background, $S_2$ liver, $S_3$ substructure inside liver (blood vessels or cancer), 
see Fig.\ref{fig:target}. The energy combines standard data term $E(S,P)$ from \eqref{eq:lk_term}, boundary length $||\partial S||$,
an inclusion constraint $S_3\subset S_2$, and a penalty for $L_2$ distance between the background segment and 
a given shape template $T$, as follows
\begin{equation} \label{eq:EDB1}
E(S,P) +\lambda||\partial S||+[S_3\subset S_2] + \beta||S_1-T||_{L2}.
\end{equation}
For fixed models $P^k$ this energy can be globally minimized over $S$ as described in \cite{DB:ICCV09}.
In this example intensity likelihood models $P^k$ are histograms  
treated as unknown parameters and estimated using block-coordinate descent for variables $S$ and $P$. 
Figure \ref{fig:target} compares optimization of \eqref{eq:EDB1} in (b)
with optimization of a modified energy replacing standard likelihoods $E(S,P)$ with a weighted
data term in \eqref{eq:wlk_term} 
\begin{equation} \label{eq:EDB2}
E_W(S,P) +\lambda||\partial S||+[S_3\subset S_2] + \beta||S_1-T||_{L2}
\end{equation}
for fixed weights $W$ set from specific target volumes (c-d). 

The teaser in Figure \ref{fig:teaser}(c) demonstrates a similar example for separating a kidney 
from a liver based on Gaussian models $P^k$, as in Chan-Vese \cite{Chan-Vese-01b}, instead of histograms. 
Standard likelihoods $E(P,S)$ in \eqref{eq:EDB1} show equal-size bias, which is corrected by weighted likelihoods $E_W(P,S)$
in \eqref{eq:EDB2} with approximate target volumes $W=\{0.05,0.95\}$.

\subsection{Segmentation without volumetric bias} \label{sec:sgm_no_bias}

We demonstrate in different applications a practically significant effect of removing the volumetric bias, i.e., using our 
functional $\hat E(S,P)$. We first report comprehensive comparisons of binary segmentations on the GrabCut data set \cite{GrabCuts:SIGGRAPH04}, which 
consists of $50$ color images with ground-truth segmentations and user-provided bounding boxes\footnote{http://research.microsoft.com/en-us/um/cambridge/projects\\ /visionimagevideoediting/segmentation/grabcut.htm}. We compared three energies: high-order energy $\hat{E}(S,P)$ \eqref{eq:E2_H}, standard likelihoods
 $E(S,P)$ \eqref{eq:lk_term}, which was used in the well-known GrabCut algorithm \cite{GrabCuts:SIGGRAPH04}, and $E_W(S,P)$ \eqref{eq:wlk_term}, which 
constrains the solution with true target volumes (i.e., those computed from ground truth). The appearance models in each energy were based on 
histograms encoded by $16$ bins per channel, and the image data is based color specified in RGB coordinates. For each energy, we added a standard contrast-sensitive
regularization term \cite{GrabCuts:SIGGRAPH04,BJ:01}: $\lambda \sum_{p,q \in {\cal N}} \alpha_{pq} [S_p \neq Sq]$, where $\alpha_{pq}$ denote standard pairwise 
weights determined by color contrast and spatial distance between neighboring pixels $p$ and $q$ \cite{GrabCuts:SIGGRAPH04,BJ:01}. ${\cal N}$ is the set neighboring 
pixels in a 8-connected grid. 

We further evaluated two different optimization schemes for high-order energy $\hat E(S,P)$: (i) bound optimization and (ii) high-order optimization
of concave cardinality potential $H(S)$ using polygonal approximations; see Sec.\ref{seg:energy_formulation} for details. Each energy is optimized by alternating two iterative steps: (i) fixing the appearance histogram models and optimizing the energy w.r.t $S$ using graph cut \cite{BK:PAMI04}; and (ii) fixing segmentation $S$ and updating the histograms from current solution. For all methods we used the same appearance model initialization based on a user-provided box\footnote{The data set comes with two boxes enclosing the foreground segment for each image. We used the outer bounding box to restrict the image domain and the inner box to compute initial appearance models.}. 

The error is evaluated as the percentage of mis-classified pixels with respect to the ground truth. Table \ref{tab:GrabCut} reports the best average error over $\lambda \in [1 \dots 30]$ for each method. As expected, using the true target volumes yields the lowest error. The second best performance was obtained by $\hat E(S,P)$ with high-order optimization; {\em removing the volumetric bias substantially improves the performance of standards log-likelihoods reducing the error by $6\%$}. The bound optimization obtains only a small improvement as it is more likely to get stuck in weak local minima. We further show representative examples for $\lambda=16$ in the last two rows of Table \ref{tab:GrabCut}, which illustrate clearly the effect of both equal-size bias in \eqref{eq:lk_term} and the corrections we proposed in \eqref{eq:E2_H} and \eqref{eq:wlk_term}. 

It is worth noting that the error we obtained for standard likelihoods (the last column in Table \ref{tab:GrabCut}) 
is significantly higher than the $8\%$ error previously reported in the literature, e.g., \cite{Vicente2009}. 
The lower error in \cite{Vicente2009} is based on a different (more recent) set of tighter bounding boxes \cite{Vicente2009}, 
where the size of the ground-truth segment is roughly half the size of the box.
Therefore, the equal-size bias in $\hat{E}(S,P)$ \eqref{eq:E2_H} for this particular set of boxes has an effect similar to
the effect of true target volumes $W$ in $E_W(S,P)$ \eqref{eq:wlk_term} (the first column in Table \ref{tab:GrabCut}), which 
significantly improves the performance of standard likelihoods (the last column). 
In practice, both 50/50 boxes and true $W$ are equally unrealistic assumptions that require knowledge of the ground truth.  

\begin{table*}[th]
\begin{center}
\begin{tabular}{|l||c|c|c|c|}
\hline
\hspace{0.25cm} 
 {\bf Energy} &  \parbox{22ex}{ \vspace{0.1cm} $E_W(S,P)\quad$   \eqref{eq:wlk_term} \\ \emph{true target volumes} $W$  } & 
\parbox{22ex}{$\hat{E}(S,P)\quad$ \eqref{eq:E2_H}  \\ \emph{high-order optimization}} &   \parbox{22ex}{$\hat{E}(S,P)\quad$ \eqref{eq:E2} \\ \emph{bound optimization}} 
& \parbox{22ex}{ $E(S,P)\quad$  \eqref{eq:lk_term} \\ \emph{standard likelihoods}} \\ 
\hline \hline  &  \\   [-2.65ex] \parbox{22ex} {$\mbox{}$ \\ {\bf Overall Error}   \\ $\mbox{~~}$ (50 images)}  \hspace{-1.5cm} 
& $ {\bf 5.29} \%$  & ${\bf 7.87} \%$ &  ${\bf 13.41} \%$ & ${\bf 14.41} \%$ \\ \hline  & \\ [-2.65ex]   \hspace{0.25cm}  {\bf Examples} & 
\parbox{22ex} {$\mbox{}$ \\  \\ \begin{minipage}{.3\textwidth}  \includegraphics[width=.5\linewidth]{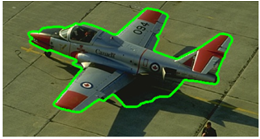} 
    \end{minipage}  \\ {\small error: $4.75\%$} \\ $\mbox{}$ \\ 
\begin{minipage}{.3\textwidth}  \includegraphics[width=.5\linewidth]{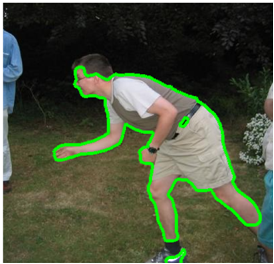}   \end{minipage} \\ 
{\small  error: $2.29\%$} \\} 
\hspace{-1cm}  & 
\parbox{22ex} { $\mbox{}$ \\ \begin{minipage}{.3\textwidth}  \includegraphics[width=.5\linewidth]{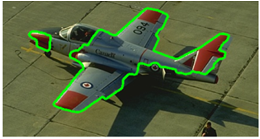} 
    \end{minipage} \\   {\small error: $6.85\%$} \\ $\mbox{}$ \\  \begin{minipage}{.3\textwidth}  \includegraphics[width=.5\linewidth]{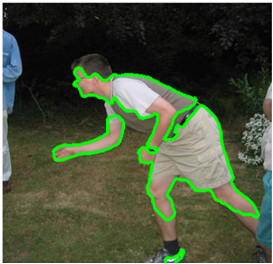} 
    \end{minipage} \\  {\small error: $4.95\%$}} \hspace{-1cm} &   
\parbox{22ex} {\begin{minipage}{.3\textwidth}  \includegraphics[width=.5\linewidth]{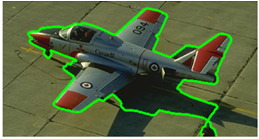} 
    \end{minipage} \\ {\small error: $9.64\%$} \\ $\mbox{}$ \\ \begin{minipage}{.3\textwidth}  \includegraphics[width=.5\linewidth]{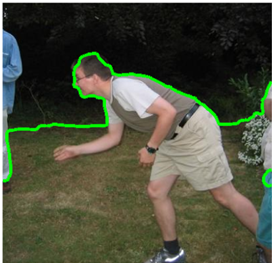} 
    \end{minipage}\\ {\small error: $41.20\%$ }} \hspace{-1cm}
   & \parbox{22ex} {\begin{minipage}{.3\textwidth}
      \includegraphics[width=.5\linewidth]{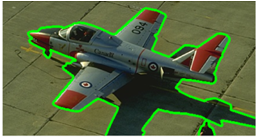} 
 \end{minipage} \\  {\small error: $14.69\%$} \\  $\mbox{}$ \\ \begin{minipage}{.3\textwidth}
      \includegraphics[width=.5\linewidth]{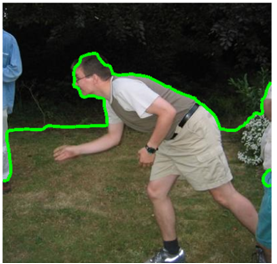} 
 \end{minipage} \\ {\small error: $40.88\%$}}  \hspace{-1cm}
 \\ \hline
\end{tabular}
\end{center}
\vspace{-2mm}
\caption{%
    Comparisons on the GrabCut data set.}
\label{tab:GrabCut}
\end{table*}

\begin{figure*}[th]
\begin{center}
\begin{tabular}{cccc}
 & \hspace{-0.45cm}  $\gamma<1$   & \hspace{-0.45cm} $\gamma=1$  & \hspace{-0.45cm} $\gamma>1$ \\
image &  \hspace{-0.45cm}  {\bf {\em equality} bias}  &  \hspace{-0.45cm} {\bf no bias} & \hspace{-0.45cm} {\bf {\em disparity} bias}\\
\includegraphics[height=1.5in]{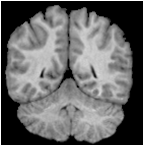} &  \hspace{-0.45cm}
\includegraphics[height=1.5in]{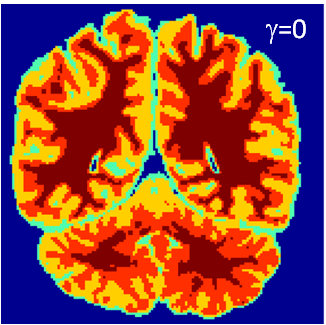} &
 \hspace{-0.45cm} 
\includegraphics[height=1.5in]{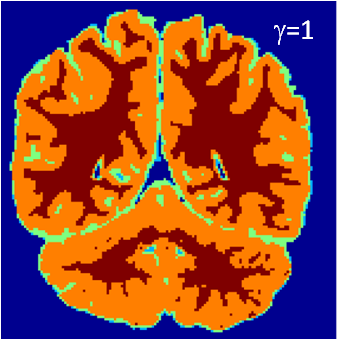} & 
\hspace{-0.45cm}
\includegraphics[height=1.5in]{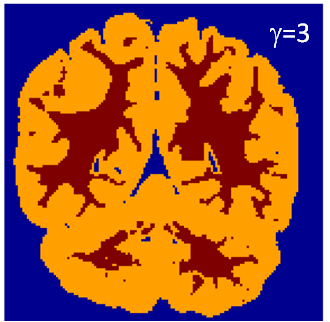}\\
&  \hspace{-0.45cm}
\includegraphics[width=.2\linewidth]{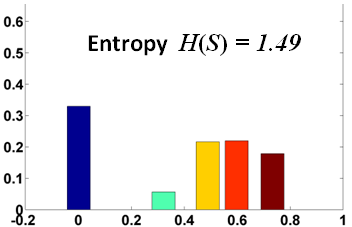}  & \hspace{-0.45cm} 
\includegraphics[width=.2\linewidth]{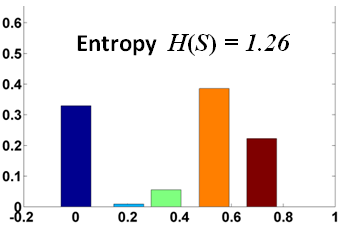}   & 
\hspace{-0.45cm}
\includegraphics[width=.2\linewidth]{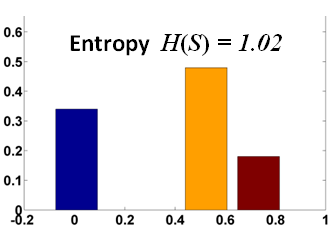} \\
& (a) & (b) & (c)
\end{tabular}
\end{center}
\caption{Segmentation using energy \eqref{eq:E2_Hgamma} combined with a standard total-variation regularization \cite{Yuan2010}. We used the Chan-Vese model \cite{Chan-Vese-01b} as appearance term and bound optimization to compute a local minimum of the energy (See Fig. \ref{fig:bound}). At each iteration, the bound is optimized w.r.t segmentation using the convex-relaxation technique in \cite{Yuan2010}. Initial number of models: $5$. $\lambda=0.1$, $\sigma=0.05$. Upper row (from left to right): image data and the results for $\gamma=0$, $\gamma=1$ and $\gamma=3$. Lower row: histograms of the number of assignments 
to each label and the entropies obtained at convergence.
\label{fig:Brain_Convex_Relaxation}}
\end{figure*}

Fig. \ref{fig:Brain_Convex_Relaxation} depicts a different application, where we segment a magnetic resonance image (MRI)
of the brain into multiple regions ($K>2$). Here we introduce an extension of $\hat{E}(S,P)$ using a positive factor $\gamma$ 
that weighs the contribution of entropy against the other terms:     
\begin{equation}
\hat{E}_\gamma(S,P) = E(S,P) + \gamma |\Omega| H(S). \label{eq:E2_Hgamma}
\end{equation}
This energy could be written as
\begin{equation*}
\sum_{k=1}^{K} |S^k| KL(I^k|P^k) + |\Omega|H(S|I) + (\gamma -1)|\Omega| H(S))
\end{equation*}
using the high-order decomposition of likelihoods $E(S,P)$ from \cite{UAI:97} presented in the intro.
Thus, the bias introduced by $H(S)$ has two cases: $\gamma\leq 1$ (volumetric {\em equality} bias) 
and $\gamma\geq 1$ (volumetric {\em disparity} bias), as discussed below. 

We used the Chan-Vese data term \cite{Chan-Vese-01b}, which assumes the appearance models in $E(S,P)$ are Gaussian distributions: $-\log P^k(I_p) \utc (I_p - \mu^k)^2/2\sigma^2$, 
with $\mu^k$ the mean of intensities within segment $S^k$ and $\sigma$ is fixed for all segments. We further added a standard total-variation term \cite{Yuan2010} 
that encourages boundary smoothness. 

The solution is sought following the bound optimization strategy we discussed earlier; See Fig. \ref{fig:bound}. The algorithm alternates between two iterative steps: (i) optimizing a bound of $\hat{E}_\gamma(S,P)$ w.r.t segmentation $S$ via a continuous convex-relaxation technique \cite{Yuan2010} while model parameters are fixed, and (ii) fix segmentation $S$ and update parameters $\mu^k$ and $w^k$ using current solution. We set the initial number of models to $5$ and fixed $\lambda=0.1$ and $\sigma=0.05$. We run the method for $\gamma=0$, $\gamma=1$ and $\gamma=3$. Fig. \ref{fig:Brain_Convex_Relaxation} displays the results using colors encoded by the region means obtained at convergence. Column (a) demonstrates the equal-size bias for $\gamma=0$; notice that the yellow, red and brown components have approximately 
the same size. Setting $\gamma=1$ in (b) removed this bias, yielding much larger discrepancies in size between these components. In (c) we show that using large 
weight $\gamma$ in energy \eqref{eq:E2_Hgamma} has a sparsity effect; it reduced the number of distinct segments/labels from $5$ to $3$. At the same time, for $\gamma > 1$, this
energy introduces {\em disparity} bias; notice the gap between the volumes of orange and brown segments has increased compared to $\gamma=1$ in (b), where there 
was no volumetric bias. This disparity bias is opposite to the equality bias for $\gamma < 1$ in (a).        

\subsection{Geometric model fitting} \label{sec:stereo}

Energy minimization methods for geometric model fitting problems have recently gained popularity due to \cite{PEARL:IJCV12}.
Similarly to segmentation these methods are often driven by a maximum likelihood based data term measuring model fit to the particular feature.
The theory presented in Section \ref{seg:energy_formulation} applies to these problems as well and they therefore exhibit the same kind of volumetric bias.

\begin{figure}[htb]
\begin{center}
\includegraphics[width=70mm]{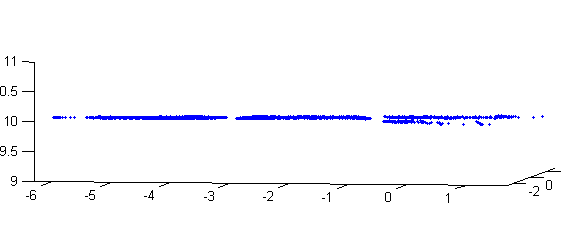}
\end{center}
\caption{3D-Geometry of the book scene in Figure~\ref{fig:teaser} (b).}
\label{fig:homographygeometry}
\end{figure}

Figures~\ref{fig:teaser} (b) shows a simple homography estimation example.
Here we captured two images of a scene with two planes and tried to fit homographies to these (the right image with results is shown in Figure~\ref{fig:teaser}). 
For this image pair SIFT \cite{sift:04} generated 3376 matches on the larger plane (paper and floor) and 135 matches on the smaller plane (book).
For a pair of matching points $I_p = \{x_p,y_p\}$ we use the log likelihood costs
\begin{equation}
\sum_{p\in S^k} - \log \left(w^k \cdot P^{H_k,\Sigma_k}(I_p)\right),
\end{equation}
where $P^{H_k,\Sigma_k}(I_p) = \frac{1}{(2 \pi)^2 \sqrt{|\Sigma_k|}} e^{-\frac{1}{2}d_{H_k,\Sigma_k}(x_p,y_p)^2}$ and 
$d_{H_k,\Sigma_k}$ is the symmetric mahalanobis transfer distance.
The solution to the left in Figure~\ref{fig:teaser} (b) was generated by optimizing over homographies and covariances while keeping 
the priors fixed and equal ($w^1 = w^2 = 0.5$). The volume bias makes the smaller plane (blue points) grab points from the larger plane.
For comparison Figure~\ref{fig:teaser} (b) also shows the result obtained when reestimating $w^1$ and $w^2$. Note that the two algorithms were started with the same homographies and covariances. Figure~\ref{fig:homographygeometry} shows an independently computed 3D reconstruction using the same matches as for the homography experiment.

\subsubsection{Multi Model Fitting}
Recently discrete energy minimization formulations have been shown to be effective for geometric model fitting tasks \cite{PEARL:IJCV12,LabelCosts:IJCV12}. These methods effectively handle regularization terms 
needed to produce visually appealing results. The typical objective functions are of the type
\begin{equation}
E(S,W,\Theta) = V(S)+D(S,W,\Theta)+L(S),
\label{eq:modelfittingenergy}
\end{equation}
where $V(S) = \sum_{(p,g)\in \mathcal{N}} V_{pq}(S_p,S_q)$ is a smoothness term and $L(S)$ is a label cost preventing 
over fitting by penalizing the number of labels. The data term
\begin{equation}
D(S,W,\Theta) = -\sum_k \sum_{S_p=k}\left( \log(w^k) + P(m_p|\Theta^k)\right)
\label{eq:modelfittingdataterm}
\end{equation}
consists of log-likelihoods for the observed measurements $m_p$, given the 
model parameters $\Theta$.
Typically the prior distributions $w^k$ are ignored (which is equivalent to 
letting all $w^k$ be equal) hence resulting in a bias to equal partitioning. 
Because of the smoothness and label cost terms the bias is not as evident in 
practical model fitting applications as in k-means, but as we shall see it is still present. 

Multi model fitting with variable priors presents an additional challenge.
The PEARL (Propose, Expand And Reestimate Labels) paradigm \cite{PEARL:IJCV12}  
naturally introduces and removes models during optimization.
However, when reestimating priors, a model $k$ that is not in the current labeling 
will have $w^k=0$ giving an infinite log-likelihood penalty. Therefore a simple alternating approach 
(see {\em bound optimization} in Sec.\ref{seg:energy_formulation}) will be unable to add new models to the solution. 
For sets of small cardinality it can further be seen that the entropy bound in Figure~\ref{fig:bound} 
will become prohibitively large since the derivative of the entropy function is unbounded (when approaching $w=0$).
Instead we use $\alpha$-expansion moves with higher order interactions to handle the entropy term, as described in Section~\ref{seg:energy_formulation}.

\begin{figure}[t!]
\begin{center}
\begin{tabular}{cc}
(a) & (b) \\
\includegraphics[width=40mm]{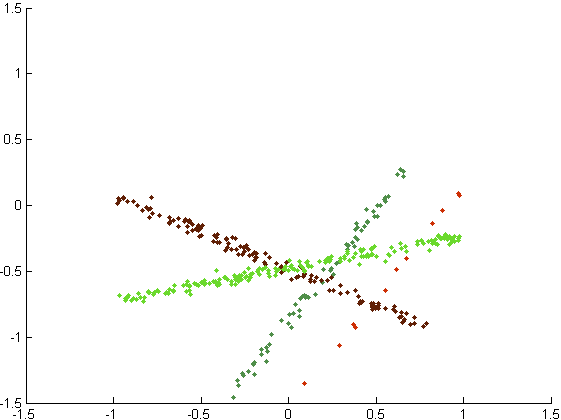} &
\includegraphics[width=40mm]{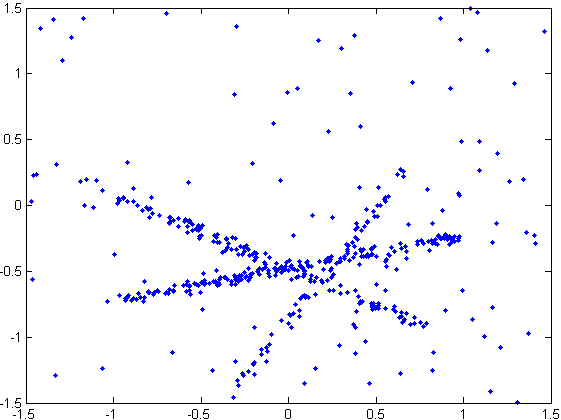}\\
(c) & (d) \\
\includegraphics[width=40mm]{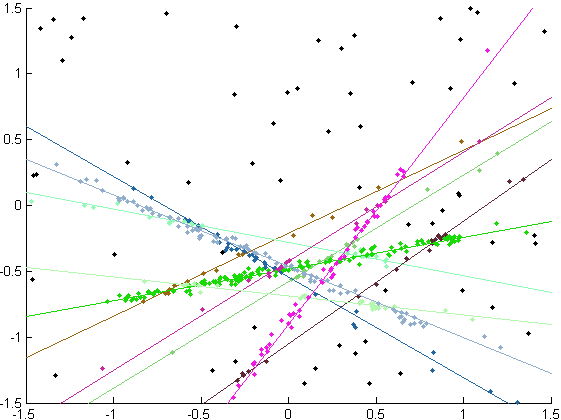} &
\includegraphics[width=40mm]{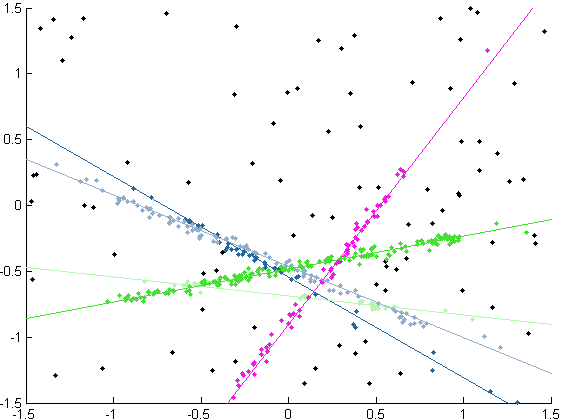} \\
(e) & (f) \\
\includegraphics[width=40mm]{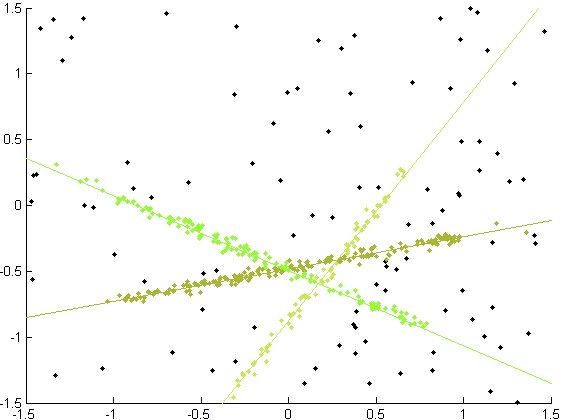} &
\includegraphics[width=40mm]{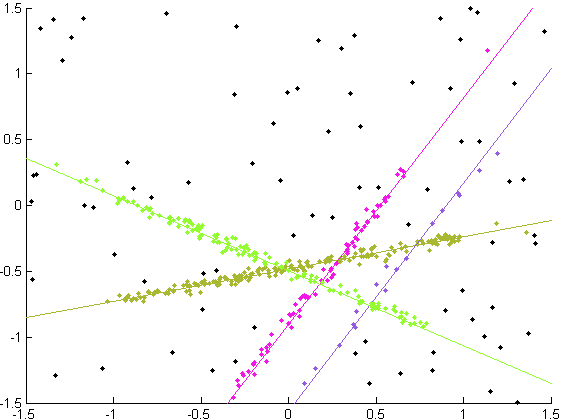}
\end{tabular}
\end{center}
\caption{{\em Line fitting:} (a) data generated from three lines, (b) data with outliers, (c) fixed $W$ and $h=100$,
(d) fixed $W$ and $h=200$, (e) fixed $W$ and $h=300$, (f) variable $W$ and $h=5$.
\label{fig:linefittingresult}} \vspace{-1ex}
\end{figure}

Figure~\ref{fig:linefittingresult} shows the result of a synthetic line fitting experiment.
Here we randomly sampled points from four lines with different probabilities, added noise with $\sigma = 0.025$ and added outliers.
We used energy \eqref{eq:modelfittingenergy} without smoothness and with label cost $h$ 
times the number of labels (excluding the outlier label). 
The model parameters $\Theta$ consist of line location and orientation. 
We treated the noise level for each line as known. 
Although the volume bias seems to manifest itself more clearly when the variance is reestimated, 
it is also present when only the means are estimated.
\begin{figure}[t!]
\begin{center}
\begin{center}
\includegraphics[width=40mm]{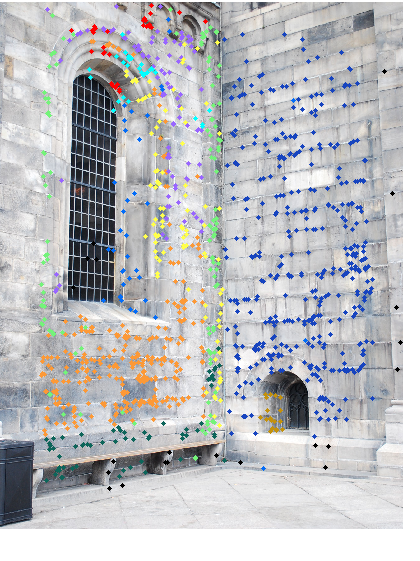}
\includegraphics[width=40mm]{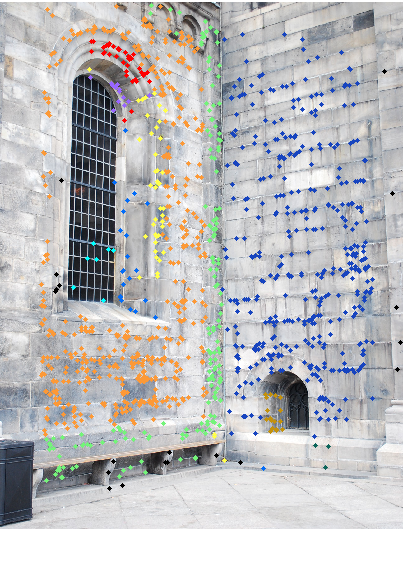}
\end{center}
\caption{{\em Homography fitting}: fixed (\emph{left}) and variable $W$ (\emph{right}).}
\label{fig:cornersolutions}
\begin{center}
\includegraphics[width=40mm]{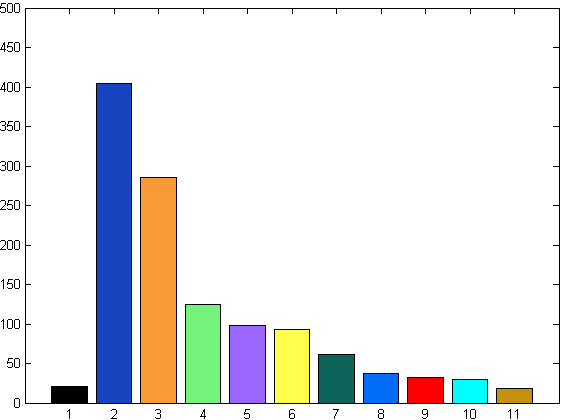}
\includegraphics[width=40mm]{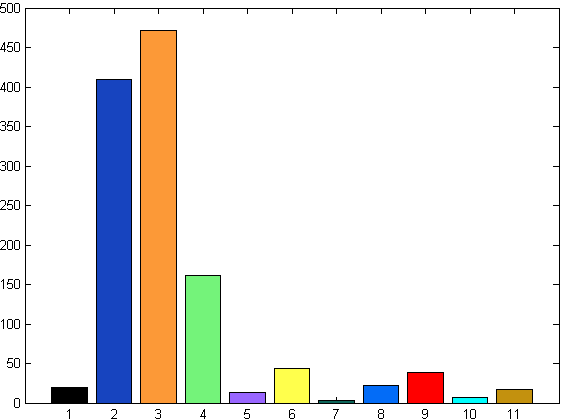}
\end{center}
\caption{Histogram of the number of assignments to each label (model) in Figure \ref{fig:cornersolutions}. 
Fixed $W$ (\emph{left}) and variable $W$ (\emph{right}).
\label{fig:cornerhist}} \vspace{-1ex}
\end{center}
\end{figure}

Using random sampling we generated $200$ line proposals to be used by both methods (fixed and variable W). 
Figure~\ref{fig:linefittingresult} (c), (d) and (e) show the results with fixed W for three different strengths of label cost.
Both the label cost and the entropy term want to remove models with few assigned points.
However, the label cost does not favor any assignment when it is not strong enough to remove a model. 
Therefore it cannot counter the volume bias of the standard data term favoring more assignments to weaker models.
In the line fitting experiment of Figure~\ref{fig:linefittingresult} we varied the strength of the label cost 
(three settings shown in (c), (d) and (e)) without being able to correctly find all the 4 lines. 
Reestimation of $W$ in Figure~\ref{fig:linefittingresult} (f) resulted in a better solution.

Figures~\ref{fig:cornersolutions} and \ref{fig:cornerhist} show the results of a homography estimation problem with the smoothness term $V(S)$.
For the smoothness term we followed \cite{PEARL:IJCV12} and created edges using a Delauney triangulation with weights 
$e^{-d^2/5^2}$, where $d$ is the distance between the points.
For the label costs we used $h=100$ with fixed $W$ and $h=5$ with variable $W$.
We fixed the model variance to $5^2$ (pixels$^2$).

The two solutions are displayed in Figure~\ref{fig:cornersolutions} and Figure~\ref{fig:cornerhist} shows a histogram of the number of assigned points to each model (black corresponds to the outlier label).
Even though smoothness and label costs mask it somewhat, the bias to equal volume can be seen here as well. 

\section{Conclusions}

We demonstrated significant artifacts in standard segmentation and reconstruction methods due to bias to equal size segments
in standard likelihoods \eqref{eq:lk_term} following from the general information theoretic analysis \cite{UAI:97}. 
We proposed binary and multi-label optimization methods that either (a) remove this bias or (b) replace it by a KL divergence term for 
any given target volume distribution. Our general ideas apply to many continuous or discrete
problem formulations.

{\small
\bibliographystyle{ieee}
\bibliography{vbal_arXiv15}
}

\end{document}